\documentclass[runningheads]{llncs}
\usepackage{tabularx}
\newcolumntype{Y}{>{\centering\arraybackslash}X}
\usepackage[flushleft]{threeparttable}
\usepackage{booktabs}
\usepackage[T1]{fontenc}
\usepackage[utf8]{inputenc}
\usepackage{lmodern}
\usepackage{subcaption}
\usepackage{adjustbox}
\usepackage{amsmath, amssymb, mathtools}
\usepackage{nicefrac}

\usepackage[colorlinks=true, linkcolor=blue, citecolor=blue, urlcolor=blue]{hyperref}
\hypersetup{pdftitle={Cross-Architectural Feature Fusion for Coronary Artery Stenosis Detection in X-ray Angiography}}

\usepackage{graphicx}

\title{Cross-Architectural Feature Fusion for Coronary Artery Stenosis Detection in X-ray Angiography}
\titlerunning{Feature Fusion for Coronary Stenosis Detection in XCA}

\usepackage{float}
\usepackage{graphicx}
\begin{document}
%
%
%

\author{Haobo Zhu\inst{1}\orcidID{0009-0000-1735-3900} \and
Mojtaba Lashgari\inst{1}\orcidID{0000-0002-5307-3495} \and
Jason Chai\inst{2,3}\orcidID{0000-0002-7398-9564} \and
Giovanni Luigi De Maria\inst{2,3}\orcidID{0000-0003-3572-1855} \and
Vicente Grau\inst{1}\orcidID{0000-0001-8139-3480} \and
Abhirup Banerjee\inst{1,2}\orcidID{0000-0001-8198-5128}
}
\authorrunning{H. Zhu et al.}
%
\institute{
Institute of Biomedical Engineering, Department of Engineering Science, University of Oxford, Oxford, United Kingdom\\
\email{abhirup.banerjee@eng.ox.ac.uk} \and
Division of Cardiovascular Medicine, Radcliffe Department of Medicine, University of Oxford, Oxford, United Kingdom 
\\
\and
Oxford Heart Centre, Oxford University Hospitals NHS Trust, Oxford, United Kingdom\\}

\maketitle              
\begin{abstract}

Coronary artery disease remains a leading cause of mortality, with stenosis detection from invasive X-ray Coronary Angiography (XCA) critical for diagnosis. Manual interpretation is time-consuming, subjective, and variable, especially under low contrast and vessel overlap. Existing deep learning methods often rely solely on detection or segmentation, limiting performance.
We propose a hybrid framework, Fusion-YOLOv11-L, combining a YOLOv11-L detector with a pretrained UNet++ via a channel attention gating module. This enables fusion of semantic and vessel-aware structural features, improving localisation of stenotic regions. The UNet++ is pretrained on ARCADE vessel segmentation dataset to provide anatomically informed representations.
Experiments on the relabeled ARCADE (ARCADE*) dataset show improved performance over YOLOv11-L, achieving AP@50 of 0.585 and F1-score of 0.596. Cross-domain evaluation on a dataset from the Oxford Acute Myocardial Infarction (OxAMI) Study reveals performance degradation due to domain shift, mitigated using test-time adaptation and augmentation. Test-time augmentation improves AP@50 from 0.434 to 0.480, indicating better generalisability.
These results show that cross-architecture feature fusion and lightweight adaptation improve robustness and generalisability of automated stenosis detection in XCA.

\end{abstract}
\section{Introduction}
Coronary Artery Disease (CAD) remains the leading cause of mortality worldwide, accounting for nine million deaths annually and affecting a substantial portion
of the global population \cite{roth2020global}. The disease is primarily driven by atherosclerosis, a progressive process involving plaque accumulation within coronary arteries, leading to luminal narrowing (stenosis) and reduced myocardial perfusion \cite{knuuti20202019}.

Invasive coronary angiography, also known as X-ray Coronary Angiography (XCA), is the gold standard for visualising coronary arteries and identifying stenotic regions. During clinical procedures, clinicians visually inspect angiographic videos acquired from multiple projections to detect abnormalities \cite{lashgari2024patient}. However, this process is time-consuming and heavily dependent on operator expertise, which can lead to missed lesions or inconsistent interpretation \cite{lashgari2024patient}.

This has led to increasing interest in automated methods for detecting stenosis directly from angiographic data. Deep learning approaches have shown strong potential in image-based detection tasks, offering more consistent and objective analysis compared to manual assessment. Unlike traditional methods that rely on vessel segmentation or handcrafted features, these models learn directly from image data, making them more adaptable to complex imaging conditions.
Recent works have explored multiple strategies to improve detection performance. Danilov et al. \cite{danilov2021real} highlighted the trade-off between performance and time efficiency by exploring the performance and runtime of different detection architectures, showing that high-capacity models improve performance while lightweight models enable faster inference. Avram et al. \cite{avram2023cathai} developed a multi-stage model with a RetinaNet detector, trained on large-scale real-world data, demonstrating strong generalisation for clinical use.
Data scarcity and class imbalance remain key challenges. Seo et al. \cite{SeoSum_DiffusionBased_MICCAI2025} addressed this using diffusion-based inpainting to synthesise realistic lesions, improving performance in limited-data settings. Pascual et al. \cite{pascual2025hyperparameter} further improved YOLO-based detectors through hyperparameter optimisation, enhancing both accuracy and efficiency.
Li et al. \cite{li2025lt} used temporal features by incorporating angiographic sequences, which improved YOLO robustness to noise and variability. Molenaar et al. \cite{doi:10.1016/j.jacadv.2025.102360} demonstrated performance of a U-Net-based model on multi-centre datasets, achieving robust performance across clinical settings.
Albasrawi et al. \cite{albasrawi2026modified} improved sensitivity to subtle lesions using spatial and frequency attention, while Lalinia et al. \cite{lalinia2026stenosis} proposed a Swin UNETR-based model to capture global context efficiently.

While several existing approaches rely on detection or segmentation-based strategies for coronary stenosis localisation, there remains a lack of investigation into the explicit use of segmentation-derived features as prior information to guide and enhance the performance of detection networks.
This study explores the potential of segmentation feature fusion for stenosis detection by fusing the segmentation features from pretrained UNet++ \cite{zhou2018unetplusplus} into the backbone of a YOLOv11-L \cite{Jocher_Ultralytics_YOLO_2023}.
The key contributions can be summarised as follows: (1) a lightweight hybrid detection-segmentation framework that integrates a pretrained UNet++ features into a YOLOv11-L backbone, enabling the combination of object-level detection and vessel-aware representation learning; (2) a Channel Attention Gating (CAG) module that performs dual-source attention, allowing selective fusion of detection-driven and anatomically meaningful features; and (3) an improved cross-domain generalisability on out-of-distribution (OOD) data compared to a detection-only baseline model.


\section{Methods}
\subsection{Dataset}

The detection experiments use the relabelled ARCADE stenosis dataset (ARCADE*) \cite{Popov_2024,SeoSum_DiffusionBased_MICCAI2025}, in which stenoses with over 50\% diameter narrowing were re-annotated. The dataset comprises 1,500 XCA images (512$\times $512) from 1,500 patients, split into 1,000 training, 200 validation, and 300 test images. Stenosis severity was determined using quantitative coronary angiography, and regions with >50\% narrowing were annotated with bounding boxes. The number of localised stenoses in each split is reported in Table~\ref{tab:arcade-oxami-bbox-counts}.
In addition, to assess the generalisability of our model, we employed 48 XCA images from 27 patients of pixel resolution 512$\times $512 with locations of stenosis over 50\% from the Oxford Acute Myocardial Infarction (OxAMI) Study as a OOD test dataset. The images are at end-diastole so that the lesions have good visibility. 

\begin{table}[!th]
\centering
\footnotesize
\setlength{\tabcolsep}{6pt}
\renewcommand{\arraystretch}{1.1}
\caption{Dataset summary.}
\label{tab:arcade-oxami-bbox-counts}
\begin{tabular}{lcccc}
\hline
Dataset     & \multicolumn{3}{c}{ARCADE*}      & OxAMI \\
\cmidrule(lr){2-4} \cmidrule(lr){5-5}
Split       & Train & Validation & Test & Test \\
\hline
\# Image/ \# Stenosis & 1000/998 & 200/376 & 300/412& 48/85\\
\hline
\end{tabular}
\end{table}

\subsection{Model Architecture}
\begin{figure}[!th]
    \centering
    \includegraphics[width=0.9\linewidth]{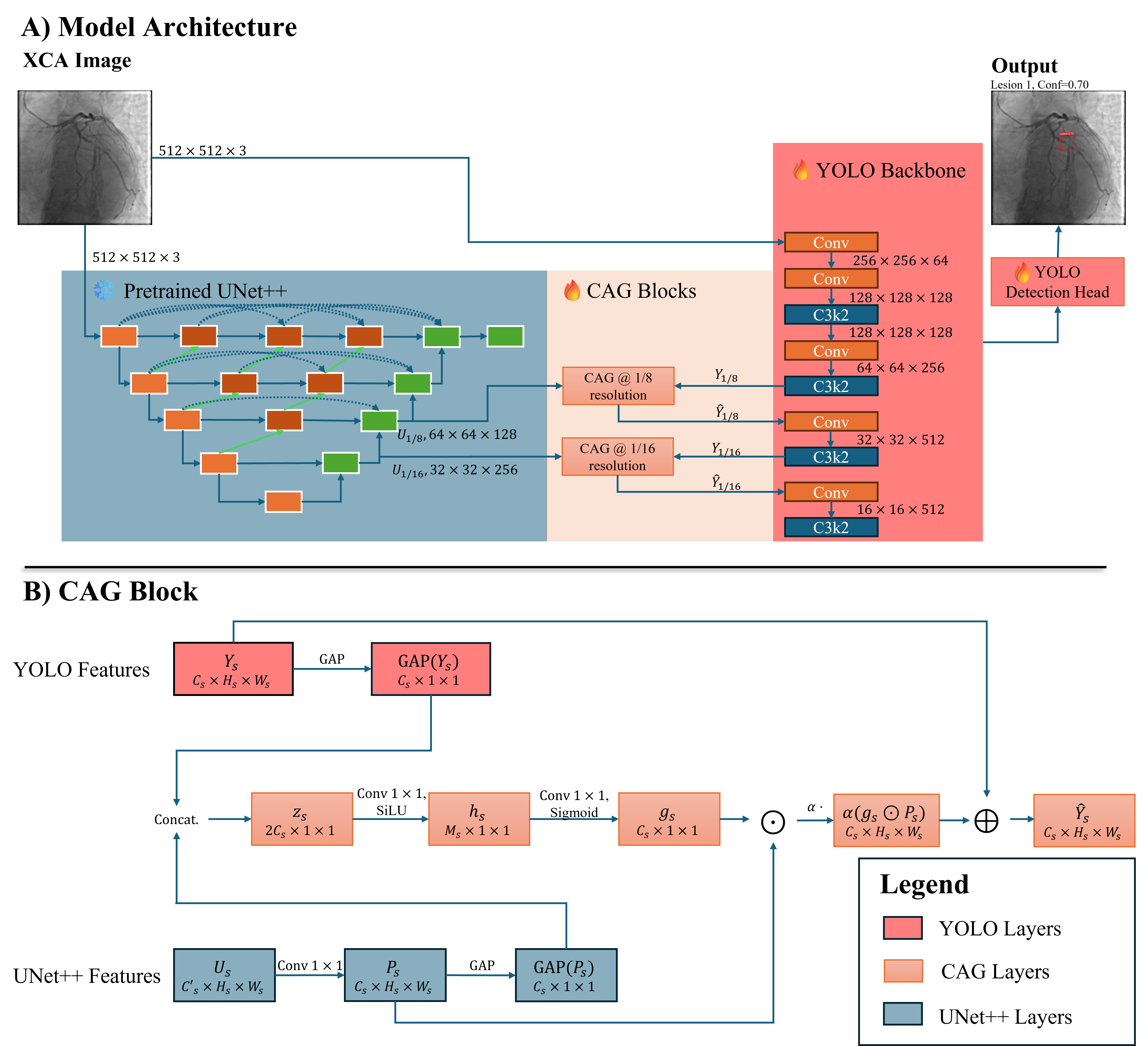}
    \caption{A) Overall model architecture; B) Channel attention gating (CAG) module for cross-architecture feature fusion between the UNet++ decoder (UNet++ branch) and the YOLOv11-L encoder (YOLO branch).}
    \label{fig:cag_module}
\end{figure}
This section presents a hybrid framework for stenosis detection in XCA images that jointly leverages object detection and vessel-aware representation learning.

To leverage the complementary strengths of a detection-optimised encoder and a segmentation-optimised representation, our framework integrates features of a pretrained UNet++ decoder~\cite{zhou2018unetplusplus} with a YOLOv11-L backbone~\cite{Jocher_Ultralytics_YOLO_2023} via a Channel Attention Gating (CAG) module, as shown in Fig.~\ref{fig:cag_module}A.
The CAG module fuses intermediate feature maps from a UNet++ decoder and a YOLOv11-L backbone at two shared scales, $\nicefrac{1}{8}$ and $\nicefrac{1}{16}$, as shown Fig.~\ref{fig:cag_module}B.
The UNet++ decoder is pretrained on the Syntax dataset of the ARCADE dataset~\cite{Popov_2024}, which provides pixel-level annotations of the coronary artery branches.
This pretraining imparts strong vessel-aware representations that complement the discriminative, detection-focused features extracted by the YOLO backbone.

\paragraph{Notation.}
Let $Y_s \in \mathbb{R}^{B \times C_s \times H_s \times W_s}$ denote the YOLOv11-L feature map at scale $s \in \{\nicefrac{1}{8}, \nicefrac{1}{16}\}$ on the encoder, and let $U_s \in \mathbb{R}^{B \times C'_s \times H_s \times W_s}$ denote the UNet++ decoder feature at the same scale, where $B$ is the batch size, $C_s$, $C'_s$ are the respective channel dimensions, and $H_s$ and $W_s$ are the height and width of the feature maps at resolution level $s$.

\paragraph{Dual-source squeeze.}
To enhance the feature maps in the YOLOv11-L backbone with the feature maps from the decoder of UNet++, we perform Global Average Pooling (GAP) squeeze~\cite{hu2018squeeze} and extension to multi-branch representations on both $Y_s$ and $P_s$, followed by channel-wise concatenation, as shown in the CAG Block depicted in Fig.~\ref{fig:cag_module}B:
\begin{equation}
    z_s = \left[\,\mathrm{GAP}(Y_s)\;\|\;\mathrm{GAP}(P_s)\,\right], \quad z_s \in \mathbb{R}^{B \times 2C_s \times 1 \times 1},
\end{equation}
where $\mathrm{GAP}(\cdot)$ denotes spatial global average pooling and $[\,\cdot\;\|\;\cdot\,]$ denotes concatenation.
This formulation enables the attention mechanism to capture complementary global statistics from both detection-driven and vessel-aware representations.

Because $C'_s \neq C_s$ in general, we first project $U_s$ into the channel space of $Y_s$ via a $1 \times 1$ convolution with learnable weight $W_{\mathrm{proj}} \in \mathbb{R}^{C_s \times C'_s \times 1 \times 1}$:
\begin{equation}
    P_s = W_{\mathrm{proj}} * U_s, \quad P_s \in \mathbb{R}^{B \times C_s \times H_s \times W_s}.
\end{equation}

\paragraph{Channel attention gating.}
As illustrated by the orange blocks in the CAG module in Fig.~\ref{fig:cag_module}, following the concatenation of $\mathrm{GAP}(Y_s)$ and $\mathrm{GAP}(P_s)$, $\mathbf{z}_s$ is passed through a bottleneck consisting of two $1 \times 1$ convolutional layers.
The first layer reduces channel dimensionality to $M_s < C_s$, followed by a Sigmoid Linear Unit (SiLU) activation~\cite{elfwing2018sigmoid}  to provide smoother gradients~\cite{khanam2024yolov11}:
\begin{equation}
    h_s = \mathrm{SiLU}\!\left(W_1 * z_s\right), \quad h_s \in \mathbb{R}^{B \times M_s \times 1 \times 1}.
\end{equation}
After the reduction of feature maps from $2C_s$ layers to $M_s$ layers, the second $1\times 1$ convolutional layer restores the dimensionality to $C_s$, and a sigmoid activation produces channel-wise attention weights:
\begin{equation}
    {g}_s = \sigma\!\left({W}_2 * {h}_s\right), \quad {g}_s \in \mathbb{R}^{B \times C_s \times 1 \times 1}.
\end{equation}

\paragraph{Feature fusion.}
As shown in blocks in the further orange blocks in the CAG module depicted in Fig.~\ref{fig:cag_module}B, the attention weights $g_s$ are applied to the projected decoder features $P_s$ via channel-wise modulation. 
The resulting gated features are combined with the original YOLO features through a residual connection:
\begin{equation}
    \hat{Y}_s = Y_s + \alpha_s \left({g}_s \odot P_s\right), \quad \hat{{Y}}_s \in \mathbb{R}^{B \times C_s \times H_s \times W_s},
\end{equation}
where $\alpha_s$ is a learnable scalar that controls the contribution of the decoder features at resolution level $s$. This design preserves the detection capability of the backbone while selectively incorporating vessel-aware contextual information. The CAG module is applied independently at each scale $s \in \{\nicefrac{1}{8}, \nicefrac{1}{16}\}$ without parameter sharing.

\subsection{Experimental Setup}
Hyperparameters are optimised using Optuna, focusing on the number of epochs, Learning Rate (LR), Bounding Box Loss Gain (BBLG), and Class Loss Gain (CLG). BBLG controls the contribution of bounding box regression to the total loss, affecting localisation accuracy, while CLG weights the classification loss, influencing stenosis detection against the background.

For each trial, the checkpoint with the highest AP@50 (Average Precision at an Intersection-over-Union (IoU) threshold of 0.5) for the validation is saved, and the final performance is reported on the test set using the configuration of this checkpoint. The hyperparameter search space is shown in Table~\ref{tab:hyperparam}, with 36 trials covering all combinations.
Standard YOLO data augmentation is applied, including HSV jitter (hue ±1.5\%, saturation ±70\%, value ±40\%), random translation (±10\%) and scaling (±50\%), horizontal flipping, and mosaic augmentation (disabled in the final 10 epochs).
Batch size is set to 16,
and weight decay is fixed at 1e-4. All other settings follow default YOLO configurations. Performance is evaluated using precision, recall, F1-score, and AP@50.
\begin{table}[!th]
\centering
\setlength{\tabcolsep}{6.5pt}
\caption{Hyperparameter search space.}
\label{tab:hyperparam}
\begin{tabular}{lcccc}
\toprule
\textbf{Hyperparameter} & \textbf{\#Epochs} & \textbf{LR} & \textbf{BBLG} & \textbf{CLG} \\
\midrule
\textbf{Search Space} & \{50, 100\} & \{2e-4, 1e-4, 5e-5\} & \{10, 7.5, 5\} & \{0.5, 0.3\} \\
\bottomrule
\end{tabular}
\end{table}

\subsection{Test-Time Adaptation and Augmentation for Cross-Domain Evaluation}


Cross-domain generalisability is evaluated by testing the model trained on ARCADE* on the OxAMI dataset. To address domain shift, Test-Time Adaptation (TTAd) is applied using Adaptive Batch Normalisation (AdaBN)~\cite{LI2018109}, where batch statistics are recalibrated on target-domain data and frozen during inference.
Test-Time Augmentation (TTAg) is also employed by performing inference on both original and horizontally flipped images, followed by aggregation of detections.
Experiments use the Fusion-YOLOv11-L checkpoint with the highest validation AP@50. Four settings are evaluated: (i) baseline inference, (ii) TTAd only, (iii) TTAg only, and (iv) TTAd + TTAg. The batch size for this experiment is set to 8.

\section{Results}
\subsection{Quantitative Results}

To provide a comprehensive evaluation, AP@50 is used as the primary metric, as it summarises performance across all confidence thresholds via the area under the precision-recall curve, unlike precision, recall, and F1-score, which depend on a fixed threshold.
AP@50 is computed for YOLOv8-L \cite{yolov8_ultralytics}, YOLOv11-L, and Fusion-YOLOv11-L using the optimum values for LR,  number of epochs, BBLG, and CLG. Table~\ref{tab:arcade-best-trials-ap50} reports the AP@50 for these models using the optimum hyperparameters for the in-domain test dataset.

\begin{table}[!th]
\centering
\caption{Comparison of  AP@50 for different models on ARCADE* test set.}
\label{tab:arcade-best-trials-ap50}
\begin{threeparttable}
\begin{tabular}{l>{\centering\arraybackslash}p{1.2cm}>{\centering\arraybackslash}p{1.2cm}>{\centering\arraybackslash}p{1.2cm}>{\centering\arraybackslash}p{1.2cm}r}
\hline
& \multicolumn{4}{c}{\textit{Optimal Hyperparameters from Optuna}} & \\
\cmidrule(lr){2-5}
Model & LR & \#Epochs & BBLG & CLG & AP@50 \\
\hline
YOLOv8-L & 1e-4 & 100 & 7.5 & 0.5 & 0.579 \\
YOLOv11-L & 1e-4 & 100 & 7.5 & 0.3 & 0.574 \\
Fusion-YOLOv11-L (Ours) & 1e-4 & 100 & 5 & 0.3 & \textbf{0.585} \\
\hline
\end{tabular}

\end{threeparttable}
\end{table}

While AP@50 provides a confidence threshold-independent evaluation, clinical deployment requires selecting a fixed operating point; therefore, precision, recall, and F1-score are analysed at different confidence thresholds.
Table \ref{tab:arcade-oxami-conf-sweep-comparison} shows the precision, recall, and F1-score at different confidence thresholds on the test set of ARCADE*, with a threshold of 0.2 giving the best F1-score (0.596). 
As the confidence threshold decreases, recall improves while precision gets worse, showing a trade-off between precision and recall.

\begin{table}[!th]
\centering
\footnotesize
\setlength{\tabcolsep}{6pt}
\renewcommand{\arraystretch}{1.1}
\caption{Evaluation of the performance of Fusion-YOLO-v11 across confidence thresholds on the ARCADE* test set.}
\label{tab:arcade-oxami-conf-sweep-comparison}
\begin{tabular}{llcccc}
\hline
 Confidence & Precision & Recall & F1-score  \\
\hline
0.250 & 0.665 & 0.529 & 0.589 \\
0.200 & 0.630 & 0.566 & \textbf{0.596} \\
0.150 & 0.564 & 0.592 & 0.578  \\
0.100 & 0.480 & 0.621 & 0.542 \\
0.050 & 0.383 & 0.682 & 0.490 \\
\hline
\end{tabular}
\end{table}

After establishing in-domain performance, the generalisability of the model is evaluated under domain shift using the OOD OxAMI dataset. To evaluate this, we use the checkpoint of Fusion-YOLOv11-L giving the best validation AP@50, comparing against the best validation AP@50 checkpoint of YOLOv11-L.
The results, presented in Table~\ref{tab:oxami-best-fusion-yolo11-tta-iou07}, show that the run with TTAg only achieves the best AP@50 (0.480).

\begin{table}[!th]
\centering
\setlength{\tabcolsep}{8pt}
\caption{Generalisability of YOLOv11-L and Fusion-YOLOv11-L across inference variants on the OxAMI dataset (AP@50).}
\label{tab:oxami-best-fusion-yolo11-tta-iou07}
\begin{threeparttable}
\begin{tabular}{l|cccc}
\hline
Model & Base & TTAd Only & TTAg Only & TTAd + TTAg \\
\hline
YOLOv11-L        & 0.339 & 0.392 & 0.383          & \textbf{0.413} \\
Fusion-YOLOv11-L & 0.434          & 0.432 & \textbf{0.480} & 0.470 \\
\hline
\end{tabular}
\end{threeparttable}
\end{table}

We further analyse the performance of Fusion-YOLOv11-L model to confidence thresholding on the OxAMI dataset. Precision, recall, and F1-score are compared for the base model without TTAd or TTAg, TTAd only, TTAg only, and with both TTAd and TTAg, at various confidence thresholds in Table~\ref{tab:oxami-sweep}.
In contrast to the in-domain evaluation (Table~\ref{tab:arcade-oxami-conf-sweep-comparison}), where the best F1-score is achieved at the confidence threshold of 0.2, the best F1-score for the cross-domain evaluation is achieved at 0.1. 

\begin{table}[!th]
\centering
\renewcommand{\arraystretch}{1.25}
\caption{Generalisability of detection performance across confidence thresholds and inference variants on the OxAMI dataset.}
\label{tab:oxami-sweep}
\begin{threeparttable}
\begin{tabularx}{\textwidth}{l| c *{2}{Y}| c *{2}{Y}| c *{2}{Y}| c *{2}{Y}}
\toprule
& \multicolumn{3}{c}{Original} & \multicolumn{3}{c}{TTAd Only} & \multicolumn{3}{c}{TTAg Only} & \multicolumn{3}{c}{TTAd + TTAg} \\
\cmidrule(lr){2-4} \cmidrule(lr){5-7} \cmidrule(lr){8-10} \cmidrule(lr){11-13}
Confidence & P & R & F1 & P & R & F1 & P & R & F1 & P & R & F1 \\
\midrule
0.250 & 0.625 & 0.235 & 0.342 & 0.658 & 0.294 & 0.407 & 0.638 & 0.353 & 0.455 & 0.612 & 0.353 & 0.448 \\
0.200 & 0.650 & 0.306 & 0.416 & 0.628 & 0.318 & 0.422 & 0.620 & 0.365 & 0.459 & 0.561 & 0.376 & 0.451 \\
0.150 & 0.622 & 0.329 & 0.431 & 0.588 & 0.353 & 0.441 & 0.574 & 0.412 & 0.479 & 0.530 & 0.412 & 0.464 \\
0.100 & 0.586 & 0.400 & \textbf{0.476} & 0.585 & 0.447 & \textbf{0.507} & 0.551 & 0.506 & \textbf{0.528} & 0.500 & 0.541 & \textbf{0.520} \\
0.050 & 0.460 & 0.471 & 0.465 & 0.467 & 0.506 & 0.486 & 0.409 & 0.553 & 0.470 & 0.384 & 0.565 & 0.457 \\
\bottomrule
\end{tabularx}
\begin{tablenotes}
    \item[*] P, R, and F1 stand for precision, recall, and F1-score, respectively.
\end{tablenotes}
\end{threeparttable}
\end{table}

\subsection{Qualitative Results}
Finally, the qualitative comparisons illustrate detection behaviour under both in-domain and cross-domain settings.
To qualitatively assess the performance of the proposed fusion model, the ground truth annotations (green bounding box) are compared with model predictions  (red bounding box) while overlaying the corresponding bounding boxes on the original images.

\begin{figure}[!th]
    \centering
    \includegraphics[width=0.8\linewidth]{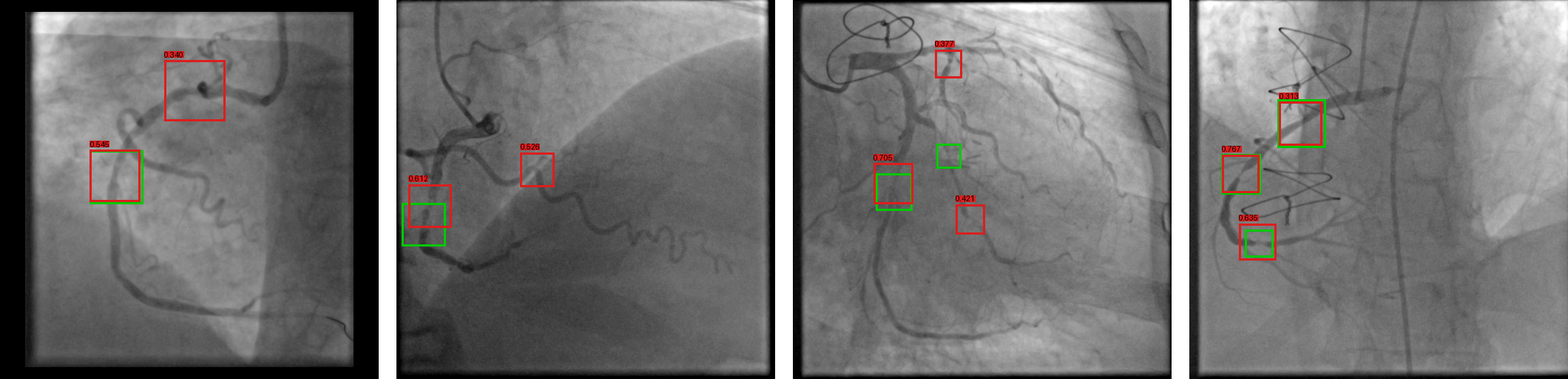}
    \caption{Comparison of ground truth (green) and predictions (red) on example XCA images from ARCADE* dataset (in-domain test dataset).}
    \label{fig:val_batch_ARCADE}
\end{figure}

\begin{figure}[!th]
    \centering

    \includegraphics[width=0.8\linewidth]{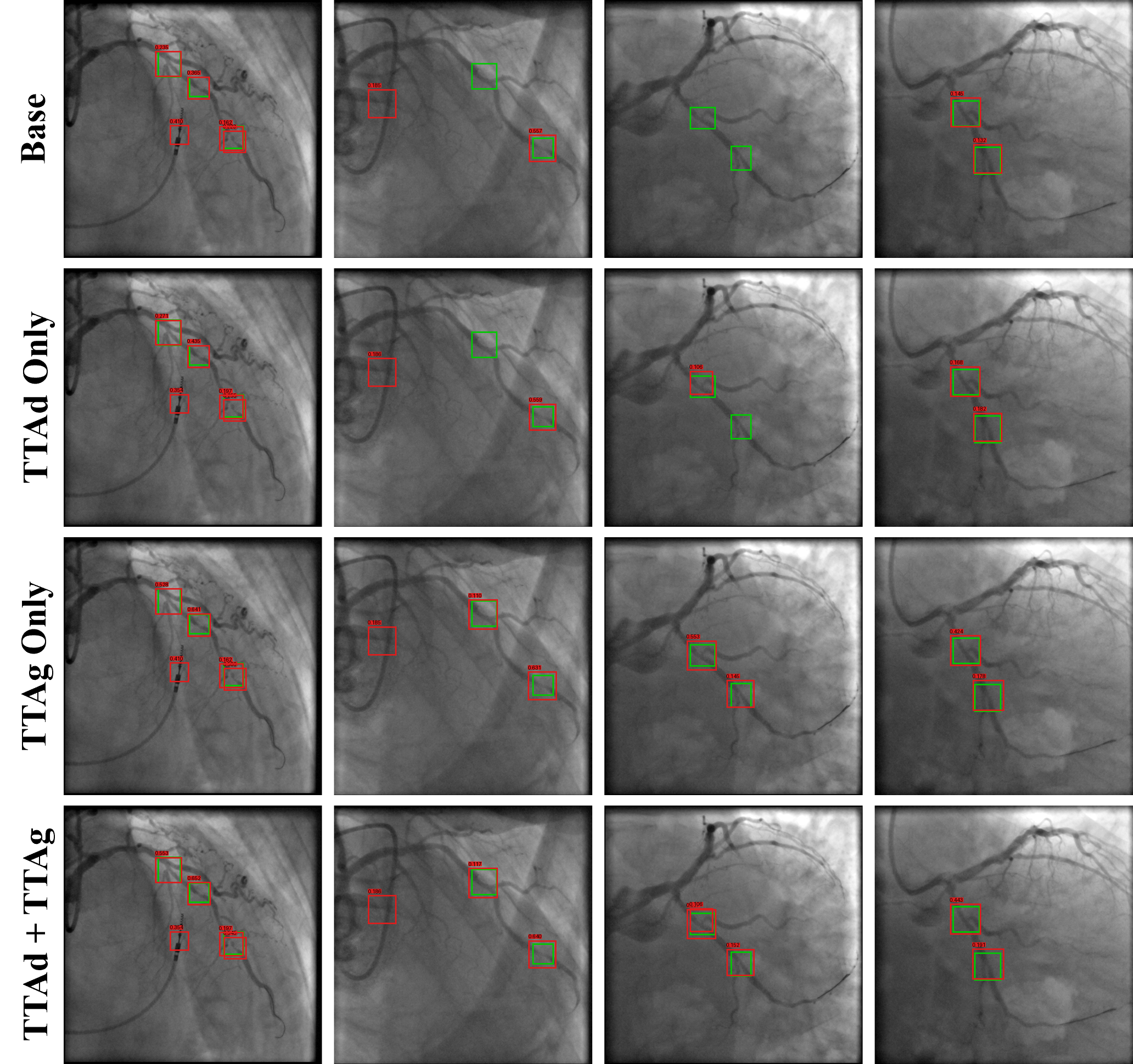}
    \caption{Comparison of ground truth (green) and predictions (red) on example XCA images from OxAMI dataset (OOD test) using Fusion-YOLOv11-L with base model without TTAd or TTAg, model with TTAd only, model with TTAg only, and model with TTAd + TTAg.}
    \label{fig:val_batch0_TTA}
\end{figure}

Figure~\ref{fig:val_batch_ARCADE} presents a comparison between the ground truth annotations from the ARCADE* dataset and the predictions produced by Fusion-YOLOv11-L with the configuration resulting in the best in-domain AP@50. Figure~\ref{fig:val_batch0_TTA} shows results on the OxAMI dataset, using the same checkpoint without TTAd or TTAg (Base), TTAd only, TTAg only, as well as TTAd + TTAg.

\section{Discussion}
The results demonstrate that both model design and hyperparameter configuration play a critical role in stenosis detection performance. The analysis based on AP@50, shown in Table~\ref{tab:arcade-best-trials-ap50}, demonstrates that incorporating feature fusion consistently improves detection performance (0.585 AP@50) compared to the baseline YOLOv11-L (0.574 AP@50), confirming the benefit of combining complementary feature representations.  
This result is consistent with our assumption that the anatomical features provided by the segmentation model can provide additional useful information for YOLO's detection backbone, boosting detection performance. While the result from YOLOv8-L (AP@50 of 0.579) is marginally better than that of the YOLOv11-L model, we have adopted YOLOv11-L as the base detection model, as it is more computationally efficient and allows better small object detection through its improved attention mechanism \cite{yolov8vsv11}.

While AP@50 provides a confidence threshold-independent evaluation, the confidence threshold analysis highlights the practical trade-off between precision and recall. As shown in Table~\ref{tab:arcade-oxami-conf-sweep-comparison}, lower confidence thresholds increase recall at the expense of precision. Notably, the best F1-score is 0.596, achieved at a confidence threshold of 0.2, indicating a balanced operating point for in-domain deployment. However, the steady increase in recall at lower thresholds suggests that the model is capable of identifying additional true positives, albeit with increased false detections. This trade-off is particularly relevant in clinical settings, where sensitivity may be prioritised depending on the application.

The cross-domain evaluation on the OxAMI dataset, shown in Table~\ref{tab:oxami-best-fusion-yolo11-tta-iou07}, reveals a noticeable performance drop compared to the in-domain ARCADE* results, reflecting the impact of domain shift in angiographic data.
While TTAd with AdaBN has not boosted the AP@50 tested on the OxAMI dataset, use of only TTAg showed the greatest improvement against the baseline.
This indicates that simple TTAg strategies can mitigate domain discrepancies without requiring retraining. 
Notably, TTAd, TTAg, and TTAd + TTAg variants all boosted the AP@50 for the baseline YOLOv11-L, but the best achievable AP@50 is significantly lower than that of Fusion-YOLOv11-L (0.413 comparing to 0.480), suggesting that the segmentation model's features play a critical role in model generalisability, making predictions more robust against the domain shift. Furthermore, the confidence threshold sweep on the OxAMI dataset, described in Table~\ref{tab:oxami-sweep}, shows that TTAg consistently improves recall and F1-score across operating points, demonstrating enhanced robustness under distribution shift.

Qualitative results in Fig. \ref{fig:val_batch0_TTA} further support these findings, showing that TTAg improves localisation when feature fusion and augmentation are applied. In particular, the model is better able to detect small or low-contrast stenotic regions, which are typically challenging in angiographic images.

Despite these improvements, several limitations remain. The model still exhibits sensitivity to confidence threshold selection, and performance degradation under domain shift highlights the need for more robust generalisation strategies. Additionally, the trade-off between precision and recall, as controlled by confidence threshold, suggests that further optimisation may be required depending on the clinical use case. Furthermore, future work could extend the proposed approach to localise stenoses directly on 3D geometry of the coronary artery tree by incorporating vessel-lumen reconstruction methods \cite{8864089}.

\section{Conclusion}
In this work, we propose a hybrid framework for coronary artery stenosis detection in x-ray angiography by fusing YOLOv11-L features with vessel-aware representations from a pretrained UNet++ decoder via a channel attention gating module. Experiments on the ARCADE* dataset show that feature fusion improves detection performance, with the best configuration achieving an AP@50 of 0.585. Confidence threshold analysis further demonstrates that the model can be tuned to different operating points, with the best F1-score of 0.596 at a threshold of 0.2.
Cross-domain evaluation on the OxAMI dataset reveals performance degradation due to domain shift; however, TTAg improves robustness, increasing AP@50 from 0.434 to 0.480 without retraining.
Overall, the results highlight the importance of feature fusion and adaptation strategies for reliable stenosis detection. Future work will explore temporal information and larger, more diverse datasets.

\begin{credits}

\subsubsection{Acknowlegements.}
The authors acknowledge the use of the services/facilities of the Institute of Biomedical Engineering (IBME), Department of Engineering Science, University of Oxford.
The work of AB was supported by the Royal Society University Research Fellowship (grant no. URF{\textbackslash}R1{\textbackslash}221314) and the British Heart Foundation (BHF) Oxford Centre of Research Excellence (RE/24/130024).
The work of ML was supported by the Royal Society Enhanced Research Expenses Grant, awarded to AB.
The OxAMI study was supported by BHF grant (CH/16/1/32013), BHF Oxford Centre of Research Excellence (RG/13/1/30181), Oxfordshire Health Services Research Committee and the National Institute for Health Research Oxford Biomedical Research Centre. 

\subsubsection{Disclosure of Interests.}
The authors have no competing interests.
\end{credits}
\bibliographystyle{splncs04}
\bibliography{references}

@software{Jocher_Ultralytics_YOLO_2023,
  author = {Glenn Jocher and Jing Qiu},
  title = {Ultralytics {YOLO11}},
  version = {11.0.0},
  year = {2024},
  url = {https://github.com/ultralytics/ultralytics},
  orcid = {0000-0001-5950-6979, 0000-0003-3783-7069},
  license = {AGPL-3.0}
}

@article{roth2020global,
  title={Global Burden of Cardiovascular Diseases and Risk Factors, 1990--2019: Update From the {GBD} 2019 Study},
  author={Roth, Gregory A and Mensah, George A and Johnson, Catherine O and Addolorato, Giovanni and Ammirati, Enrico and Baddour, Larry M and Barengo, No{\"e}l C and Beaton, Andrea Z and Benjamin, Emelia J and Benziger, Catherine P and others},
  journal={Journal of the American College of Cardiology},
  volume={76},
  number={25},
  pages={2982--3021},
  year={2020}, 
  publisher={Elsevier}
}

@software{yolov8_ultralytics,
  author = {Glenn Jocher and Jing Qiu and Ayush Chaurasia},
  title = {Ultralytics {YOLOv8}},
  version = {8.0.0},
  year = {2023},
  url = {https://github.com/ultralytics/ultralytics},
  license = {AGPL-3.0}
}

@misc{yolov8vsv11, title={{YOLO11} vs {YOLOv8}: A comprehensive technical comparison of real-time Vision Models}, url={https://docs.ultralytics.com/compare/yolo11-vs-yolov8#architectural-innovations}, journal={Ultralytics Docs}, author={Ultralytics}}

@article{lashgari2024patient,
  title={Patient-specific in silico {3D} coronary model in cardiac catheterisation laboratories},
  author={Lashgari, Mojtaba and Choudhury, Robin P and Banerjee, Abhirup},
  journal={Frontiers in Cardiovascular Medicine},
  volume={11},
  pages={1398290},
  year={2024},
  publisher={Frontiers Media SA}
}

@ARTICLE{8864089,
  author={Banerjee, Abhirup and Galassi, Francesca and Zacur, Ernesto and De Maria, Giovanni Luigi and Choudhury, Robin P. and Grau, Vicente},
  journal={IEEE Transactions on Medical Imaging}, 
  title={Point-Cloud Method for Automated {3D} Coronary Tree Reconstruction From Multiple Non-Simultaneous Angiographic Projections}, 
  year={2020},
  volume={39},
  number={4},
  pages={1278-1290},
}

@article{lalinia2026stenosis,
  title={Stenosis detection of coronary arteries in X-ray angiography images using Swin UNETR with self-supervised pre-training approach and PEFT strategy},
  author={Lalinia, Mehrshad and Almasganj, Farshad and Seyyedsalehi, Seyyed Ali},
  journal={Biomedical Signal Processing and Control},
  volume={112},
  pages={108617},
  year={2026},
  publisher={Elsevier}
}

@article{avram2023cathai,
  title={CathAI: fully automated coronary angiography interpretation and stenosis estimation},
  author={Avram, Robert and Olgin, Jeffrey E and Ahmed, Zeeshan and Verreault-Julien, Louis and Wan, Alvin and Barrios, Joshua and Abreau, Sean and Wan, Derek and Gonzalez, Joseph E and Tardif, Jean-Claude and others},
  journal={npj Digital Medicine},
  volume={6},
  number={1},
  pages={142},
  year={2023},
  publisher={Nature Publishing Group UK London}
}

@article{doi:10.1016/j.jacadv.2025.102360,
author = {Mitchel A. Molenaar  and Elsa Hebbo  and Jasper L. Selder  and Nikoloz Shekiladze  and Pratik B. Sandesara  and William J. Nicholson  and Folkert W. Asselbergs  and Syed Ahmad  and Daniel A. Gold  and Shaimaa M. Sakr  and Javier Oliván Bescós  and Vincent Auvray  and Martijn S. van Mourik  and Alexander Haak  and Yida Zhao  and Jelle D. Nieuwendijk  and Mark J. Schuuring  and Berto J. Bouma  and Steven A.J. Chamuleau  and Niels J. Verouden },
title = {Deep Learning–Based Segmentation of Coronary Arteries and Stenosis Detection in X-Ray Coronary Angiography},
journal = {JACC: Advances},
volume = {4},
number = {12, Part 2},
pages = {102360},
year = {2025},

}

@article{danilov2021real,
  title={Real-time coronary artery stenosis detection based on modern neural networks},
  author={Danilov, Viacheslav V and Klyshnikov, Kirill Yu and Gerget, Olga M and Kutikhin, Anton G and Ganyukov, Vladimir I and Frangi, Alejandro F and Ovcharenko, Evgeny A},
  journal={Scientific reports},
  volume={11},
  number={1},
  pages={7582},
  year={2021},
  publisher={Nature Publishing Group UK London}
}

@article{li2025lt,
  title={{LT-YOLO}: long-term temporal enhanced YOLO for stenosis detection on invasive coronary angiography},
  author={Li, Jiaxin and Tang, Xiang and Wang, Xuesong},
  journal={Frontiers in Molecular Biosciences},
  volume={12},
  pages={1558495},
  year={2025},
  publisher={Frontiers Media SA}
}

@article{albasrawi2026modified,
  title={Modified {YOLOv8x} model for coronary stenosis detection and troponin risk stratification},
  author={Albasrawi, Roaa and Ilyas, Muhammad and Althebyan, Qutaibah},
  journal={Discover Computing},
  volume={29},
  number={1},
  pages={34},
  year={2026},
  publisher={Springer}
}

@article{pascual2025hyperparameter,
  title={Hyperparameter optimization of YOLO models for invasive coronary angiography lesion detection and assessment},
  author={Pascual-Gonz{\'a}lez, Mario and Jim{\'e}nez-Partinen, Ariadna and Palomo, Esteban J and L{\'o}pez-Rubio, Ezequiel and Ortega-G{\'o}mez, Almudena},
  journal={Computers in Biology and Medicine},
  volume={196},
  pages={110697},
  year={2025},
  publisher={Elsevier}
}

@article{knuuti20202019,
  title={2019 ESC Guidelines for the diagnosis and management of chronic coronary syndromes: The Task Force for the diagnosis and management of chronic coronary syndromes of the European Society of Cardiology (ESC)},
  author={Knuuti, Juhani and Wijns, William and Saraste, Antti and Capodanno, Davide and Barbato, Emanuele and Funck-Brentano, Christian and Prescott, Eva and Storey, Robert F and Deaton, Christi and Cuisset, Thomas and others},
  journal={European heart journal},
  volume={41},
  number={3},
  pages={407--477},
  year={2020},
  publisher={Oxford University Press}
}

@article{khanam2024yolov11,
  author       = {Khanam, Rahima and Hussain, Muhammad},
  title        = {{YOLOv11}: An Overview of the Key Architectural Enhancements},
  journal      = {arXiv preprint arXiv:2410.17725},
  year         = {2024},
  eprint       = {2410.17725},
  eprinttype   = {arXiv},
  eprintclass  = {cs.CV},
}

@inproceedings{SeoSum_DiffusionBased_MICCAI2025,
  title={Diffusion-Based User-Guided Data Augmentation for Coronary
         Stenosis Detection},
  author={Seo, Sumin and Lee, In Kyu and Kim, Hyun-Woo and Min, Jaesik
          and Jung, Chung-Hwan},
  booktitle={Medical Image Computing and Computer Assisted
             Intervention -- MICCAI 2025},
  series={Lecture Notes in Computer Science},
  volume={15967},
  pages={149--159},
  publisher={Springer},
  year={2025}
}

@article{LI2018109,
title = {Adaptive Batch Normalization for practical domain adaptation},
journal = {Pattern Recognition},
volume = {80},
pages = {109-117},
year = {2018},
issn = {0031-3203},
author = {Yanghao Li and Naiyan Wang and Jianping Shi and Xiaodi Hou and Jiaying Liu}
}

@inproceedings{zhou2018unetplusplus,
  author       = {Zhou, Zongwei and Rahman Siddiquee, Md Mahfuzur and Tajbakhsh, Nima and Liang, Jianming},
  title        = {{UNet++}: A Nested {U-Net} Architecture for Medical Image Segmentation},
  booktitle    = {Deep Learning in Medical Image Analysis and Multimodal Learning for Clinical Decision Support},
  series       = {Lecture Notes in Computer Science},
  volume       = {11045},
  pages        = {3--11},
  publisher    = {Springer},
  address      = {Cham},
  year         = {2018}
}

@article{Popov_2024,
  author       = {Popov, Maxim and Amanturdieva, Akmaral and Zhaksylyk, Nuren and Alkanov, Alsabir and Saniyazbekov, Adilbek and Aimyshev, Temirgali and Ismailov, Eldar and Bulegenov, Ablay and Kuzhukeyev, Arystan and Kulanbayeva, Aizhan and Kalzhanov, Almat and Temenov, Nurzhan and Kolesnikov, Alexey and Sakhov, Orazbek and Fazli, Siamac},
  title        = {Dataset for Automatic Region-based Coronary Artery Disease Diagnostics Using {X}-Ray Angiography Images},
  journal      = {Scientific Data},
  volume       = {11},
  number       = {1},
  pages        = {20},
  year         = {2024},
}

@inproceedings{hu2018squeeze,
  author       = {Hu, Jie and Shen, Li and Sun, Gang},
  title        = {Squeeze-and-Excitation Networks},
  booktitle    = {Proceedings of the {IEEE} Conference on Computer Vision and Pattern Recognition},
  pages        = {7132--7141},
  year         = {2018}
}

@article{elfwing2018sigmoid,
  author       = {Elfwing, Stefan and Uchibe, Eiji and Doya, Kenji},
  title        = {Sigmoid-Weighted Linear Units for Neural Network Function Approximation in Reinforcement Learning},
  journal      = {Neural Networks},
  volume       = {107},
  pages        = {3--11},
  year         = {2018}
}
\end{document}